\documentclass[]{spie}  

\usepackage{amsmath,amsfonts,amssymb}
\usepackage{subcaption}
\usepackage{graphicx}
\usepackage{cite} 
\usepackage{epsfig}
\usepackage{amsmath}
\usepackage{nccmath}
\usepackage{amssymb}
\usepackage{mwe}
\usepackage{acro}
\usepackage{amssymb}
\usepackage{xcolor,colortbl}
\usepackage{tabularx}
\usepackage{relsize}
\usepackage{pifont}
\usepackage{booktabs} 
\usepackage{multirow}
\usepackage{multicol}
\usepackage{adjustbox}
\usepackage{float}
\usepackage{graphicx}
\usepackage{makecell}
\usepackage{tabu}
\usepackage[colorlinks=true, allcolors=blue]{hyperref}
\usepackage[capitalize]{cleveref}
\usepackage{algorithm}

\title{Automatic Image Unfolding and Stitching Framework for Esophageal Lining Video Based on Density-Weighted Feature Matching}   
\author[a]{Muyang Li}
\author[b]{Juming Xiong}
\author[a]{Ruining Deng}
\author[a]{Tianyuan Yao}
\author[c]{Regina N Tyree}
\author[c]{Girish Hiremath}
\author[a,b]{Yuankai Huo}

\affil[a]{Department of Computer Science, Vanderbilt University, Nashville, TN, USA}

\affil[b]{Department of Electrical and Computer Engineering, Vanderbilt University, Nashville, TN, USA}
\affil[c]{Division of Pediatric Gastroenterology, Hepatology, and Nutrition, Vanderbilt University Medical Center, Nashville, TN, USA}

\authorinfo{Corresponding author: Yuankai Huo: E-mail: yuankai.huo@vanderbilt.edu}

\begin{document} 
\maketitle

\begin{abstract}

Endoscopy is a crucial tool for diagnosing the gastrointestinal tract, but its effectiveness is often limited by a narrow field of view and the dynamic nature of the internal environment, especially in the esophagus, where complex and repetitive patterns make image stitching challenging. This paper introduces a novel automatic image unfolding and stitching framework tailored for esophageal videos captured during endoscopy. The method combines feature matching algorithms, including LoFTR, SIFT, and ORB, to create a feature filtering pool and employs a Density-Weighted Homography Optimization (DWHO) algorithm to enhance stitching accuracy. By merging consecutive frames, the framework generates a detailed panoramic view of the esophagus, enabling thorough and accurate visual analysis. Experimental results show the framework achieves low Root Mean Square Error (RMSE) and high Structural Similarity Index (SSIM) across extensive video sequences, demonstrating its potential for clinical use and improving the quality and continuity of endoscopic visual data.


\end{abstract}


\section{INTRODUCTION}
\label{sec:intro}  

Endoscopy is a prevalent medical procedure used to investigate, diagnose, and treat internal organs~\cite{8047436, valdastri2012advanced}. With advances in computer technology, it provides visible details and extents for clinical decision support and intervention~\cite{li2009targeted}. However, endoscopic imaging still has limitations that affect diagnosis and treatment~\cite{banerjee2012advances}. A major challenge is the restricted field of view provided by endoscopic cameras, limited by the insertion angle and position of the endoscope, which can prevent full observation of organs or tissues~\cite{van2014field}. This limited view can lead to missed critical areas of pathology. While endoscopes can magnify images to reveal details of lesions, magnification further narrows the field of view, obscuring the broader context needed to assess the overall condition~\cite{sano2005efficacy}. The dynamic environment, with constantly moving organs and tissues, complicates the stability and continuity of the video feed, increasing the risk of missing important diagnostic information~\cite{RN1}. These challenges are particularly severe in the esophagus, where intricate patterns and textures differ significantly from natural images. Image stitching techniques are necessary to provide a comprehensive view of internal surfaces~\cite{wang2020review}. By stitching multiple images captured during an endoscopic procedure, it is possible to create a panoramic view that offers a broader context and detailed inspection of the entire area~\cite{bergen2014stitching}. This is valuable in regions like the esophagus where intricate patterns and textures are present, enhancing diagnostic accuracy and supporting more effective treatment planning~\cite{carroll2009rectified}.

Research has focused on improving the robustness of image stitching, especially in challenging environments where traditional methods fall short. Image stitching merges multiple overlapping images to create panoramic views and relies on effective image matching techniques. Classic methods like Scale-Invariant Feature Transform (SIFT)\cite{lowe2004distinctive}, which detects and describes local features by identifying key points and computing gradients, and Oriented FAST and Rotated BRIEF (ORB)\cite{rublee2011orb}, which combines the FAST keypoint detector with a BRIEF descriptor for efficient matching, are widely used. However, these techniques often struggle in complex and dynamic scenarios, such as medical imagery, where textures are intricate and variable, particularly in the esophagus. Recent deep learning-based stitching methods, like LoFTR~\cite{2021arXiv210400680S}, a method that first establishes dense pixel-wise matches at a coarse level and then refines them using Transformer-based attention layers, have shown promise in addressing these challenges. Although these methods are still in early application to medical imagery~\cite{markova2022global, liu2022feature, farhat2023self}, they hold potential for improving stitching in environments with limited feature points and small shape variations, particularly in esophageal imaging.

\begin{figure*}[t]
\begin{center}
\includegraphics[width=0.9\linewidth]{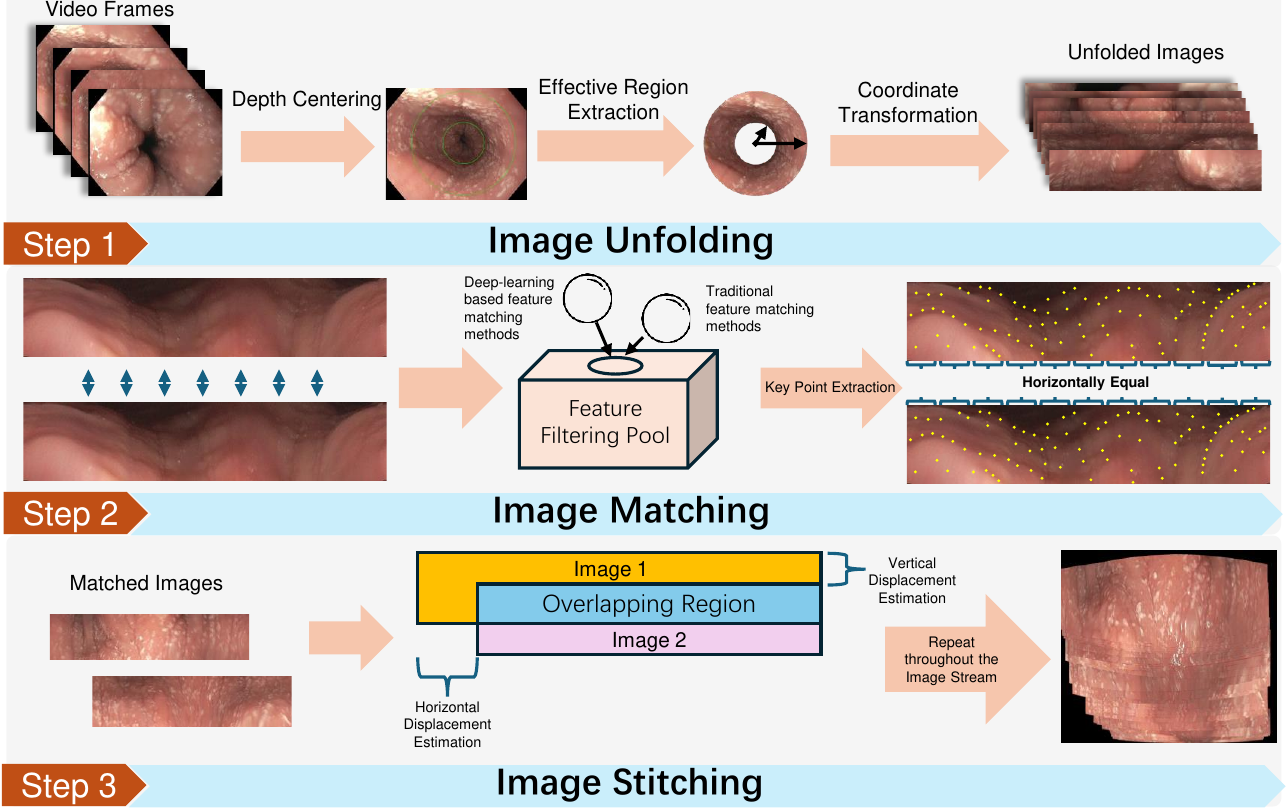}
\end{center}
\caption{Pipeline for esophageal image unfolding, matching, and stitching. The process includes three main steps: 1) transforming circular video views into unfolded images; 2) pooling and filtering feature points from deep-learning and traditional methods; 3) estimating horizontal and vertical displacements in overlapping regions to create a stitched representation of the esophageal inner surface.}
\label{fig:pipeline}
\end{figure*}

In this paper, we introduce a framework for stitching esophagus images to improve panoramic view accuracy. The process begins with a depth-based image unfolding technique using the Depth Anything model, which tracks the mean depth center for consistent image transformation, flattening the esophagus into a two-dimensional format for easier analysis. We then use feature matching with SIFT, ORB, and LoFTR to create a comprehensive feature pool, ensuring well-distributed feature points for precise stitching. Finally, our Density-Weighted Homography Optimization (DWHO) algorithm refines the stitching, resulting in a detailed panoramic view.

The contribution of this paper is threefold:
\begin{itemize}
    \item We introduce a depth-based image unfolding technique using the Depth Anything\cite{yang2024depth} model for consistent esophageal image transformation.
    \item We implement a method that controls and filters feature points by integrating LoFTR, SIFT, and ORB, selectively averaging their horizontal positions. A filtering threshold based on the depth center offset is introduced to select the most reliable feature points, ensuring consistent and parallel horizontal distribution during stitching.
    \item We propose a Density-Weighted Homography Optimization (DWHO) algorithm that integrates feature point density into homography estimation to regulate both horizontal and vertical adjustments during stitching.
\end{itemize}

\section{METHOD}
\label{sec:method}  
\subsection{Image unfolding}

\begin{figure*}[h]
\begin{center}
\includegraphics[width=1\linewidth]{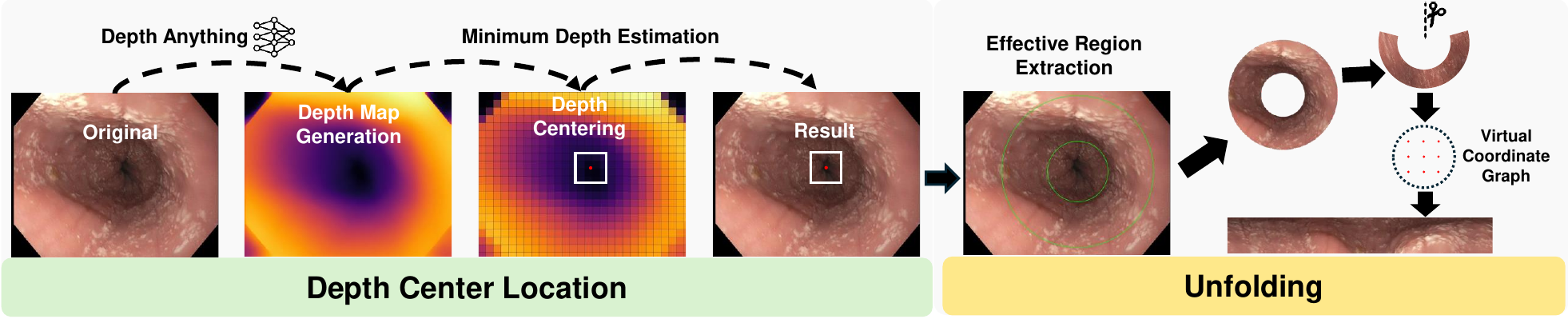}
\end{center}
\caption{Workflow of depth center location and unfolding process}
\label{fig:method}
\end{figure*}

As shown in Fig.\ref{fig:method}
, we used the Depth Anything model to process each frame of esophageal images captured by endoscopic cameras. The process begins with pre-processing steps, including resizing, normalization, and preparation for network input. The model then generates a depth map for each image, which is resized to match the original dimensions. To maintain continuity and stability, we identify the mean depth point by finding the pixel with the minimum depth value within a defined margin, excluding the sides of the image. This depth center serves as a consistent reference point across frames, crucial for reliable unfolding of the esophagus. We track the trajectory of these depth centers over time to ensure their consistency.

Next, we cut and unwrap the esophageal images to create a flat representation. The inner and outer radii ($r_{inner}$ and $r_{outer}$) are determined based on the depth centers, with $r_{outer}$ calculated as the smallest distance from each depth center to the image edge, and $r_{inner}$ set to a fixed value. To minimize distortion, we adapted a borehole video sequence unfolding method~\cite{deng2019generating}. For each image, we extract an annular region between these radii centered on the depth point and unwrap it using a Virtual Cylindrical Grid (VCG). This converts the circular region into a rectangular image, effectively ``unfolding" the esophagus into a two-dimensional format for easier analysis. This process is repeated for each frame, resulting in a sequence of unwrapped images that can be further processed or stitched together. The original images are also annotated with circles marking the $r_{inner}$ and $r_{outer}$ boundaries, providing a visual reference. The final results, including annular regions, unwrapped images, and annotated originals, are saved in organized directories for easy access and further analysis.

\subsection{Image matching}
\label{subsec}

To enhance image stitching accuracy, we implement a method that controls and filters feature points by restricting their relative distances between two images. This method integrates feature points detected by LoFTR, SIFT, and ORB, selectively averaging their horizontal positions. The innovation lies in introducing a filtering threshold based on the offset of the depth center, allowing for the selection of the most reliable feature points. By ensuring that matched points maintain consistent, parallel horizontal distribution, we reduce distortion and warping caused by feature matching errors. This approach results in a flatter, more consistent unwrapped representation of the esophagus, improving the visual quality of the images.

\subsection{Image stitching}

\subsubsection{Stitch}
\label{subsubsec}
Cylindrical projection is used to reduce distortions in image stitching, particularly for images with a wide field of view. This method maps images onto a cylindrical surface, preserving straight lines and minimizing distortion across the panorama. The process involves transforming each pixel from its original planar coordinate to a cylindrical coordinate, using intrinsic camera parameters like focal length. This transformation minimizes horizontal displacement of features, improving image alignment during stitching. By projecting images onto a cylinder, we reduce artifacts such as stretching or compressing.

The final step in image stitching involves aligning and merging processed images to form a panorama. Using the homography matrix~\cite{guiqin2019fast} estimated from matched features, each image is warped to align with the base image. The cylindrical projection ensures less distortion, while the homography matrix provides the geometric transformation needed for accurate placement. A horizontal threshold is applied during stitching to control displacement between images. If this threshold is exceeded, the stitched image is compressed horizontally to maintain consistency. A post-stitching adjustment corrects any minor misalignments, further compressing the stitched image if necessary to ensure the final panorama is free from noticeable artifacts or distortions.

\subsubsection{DWHO Algorithm}

Our proposed Density-Weighted Homography Optimization (DWHO) algorithm is an enhancement of the LHOO algorithm~\cite{deng2023geological}, originally designed for vertical borehole systems. Given the significant horizontal displacement and unpredictable distortions in esophagus video sequences, our method introduces additional control and evaluation in the horizontal direction. The DWHO algorithm integrates feature points detected using multiple feature matching methods, including LoFTR, SIFT, and ORB. Initial homography matrices between consecutive images are estimated using the MSAC (M-estimator SAmple Consensus)~\cite{zhang2022hyperspectral} algorithm, which accommodates the presence of outliers. Confidence values are calculated based on the inliers identified by MSAC, and density weights are assigned based on the horizontal distribution of feature points, preventing regions with high density from disproportionately influencing the stitching process.

A critical aspect of our approach is the calculation of horizontal displacement (\( D \)), which addresses the horizontal shifts inherent in esophagus video sequences. \( D \) is computed as the weighted average of the horizontal distances between matched feature points, with the formula:

\[
D = \frac{\sum_{i=1}^{n} d_i \cdot w_i}{\sum_{i=1}^{n} w_i}
\]

where \( d_i \) represents the horizontal distance between corresponding feature points, and \( w_i \) is the associated density weight. After calculating \( D \), a vertical offset threshold is determined, and an optimal vertical offset is calculated to minimize the weighted difference between homography elements and the threshold. Finally, an iterative process is used to apply these optimal parameters across the image sequence, yielding a set of stitching parameters that enable seamless image alignment.

\begin{algorithm}[H]
\caption{DWHO Algorithm}
\label{alg:dwho}
\textbf{Input:} Image sequence \{F(1), F(2), \dots, F(N)\}, Feature Points from LoFTR, SIFT, and ORB, N is total frame number\\
\textbf{Output:} Stitching parameters \{O(1), O(2), \dots, O(N)\}

\begin{enumerate}\setlength{\itemsep}{0pt}  
    \item \textbf{Obtain:} Homography matrices \(H^1, H^2, \dots, H^{N-1}\) using MSAC.
    \item \textbf{Calculate:} Density weight \(w_i\).
    \item \textbf{Compute:} Horizontal Displacement \(D = \frac{\sum_{i=1}^{n} d_i \cdot w_i}{\sum_{i=1}^{n} w_i}\).
    \item \textbf{Compute:} Optimal offset \(O(i) = \min \left\{ w_i \cdot \left| H_{i} - \left[\epsilon - 3(\epsilon - c_i)\right] \right| + D \right\}\).
    \item \textbf{Iterate:} Update \(O(i)\) for each frame.
    \item \textbf{Output:} Data set \{O(1), O(2), \dots, O(N)\}.
\end{enumerate}
\end{algorithm}

\section{Experiments}
\label{sec:data}
\subsection{Data}
In this study, five videos were collected from five children aged 6-18 with Eosinophilic Esophagitis (EoE) at Monroe Carrell Jr. Children’s Hospital at Vanderbilt University Medical Center (VUMC). The study protocol, which does not involve any procedures beyond routine care, was approved by the Vanderbilt University Medical Center’s Institutional Review Board. Participants were enrolled after obtaining appropriate consent from the caregiver and assent from the child.

\subsection{Evaluation metrics}
\begin{itemize}
    \item \textbf{SSIM}
    Structural Similarity Index (SSIM) is a metric used to measure the similarity between two images. Unlike traditional methods like Mean Squared Error (MSE) or Root Mean Squared Error (RMSE), which only consider pixel-by-pixel differences, SSIM considers changes in structural information, luminance, and contrast.
    \item \textbf{RMSE}
    Root Mean Squared Error (RMSE) is a standard way to measure the error between predicted values and the actual values. It is the square root of the average of the squared differences between the predicted and actual values.
\end{itemize}

\section{RESULTS}
\label{sec:result}  
\begin{figure*}[t]
\begin{center}
\includegraphics[width=0.9\linewidth]{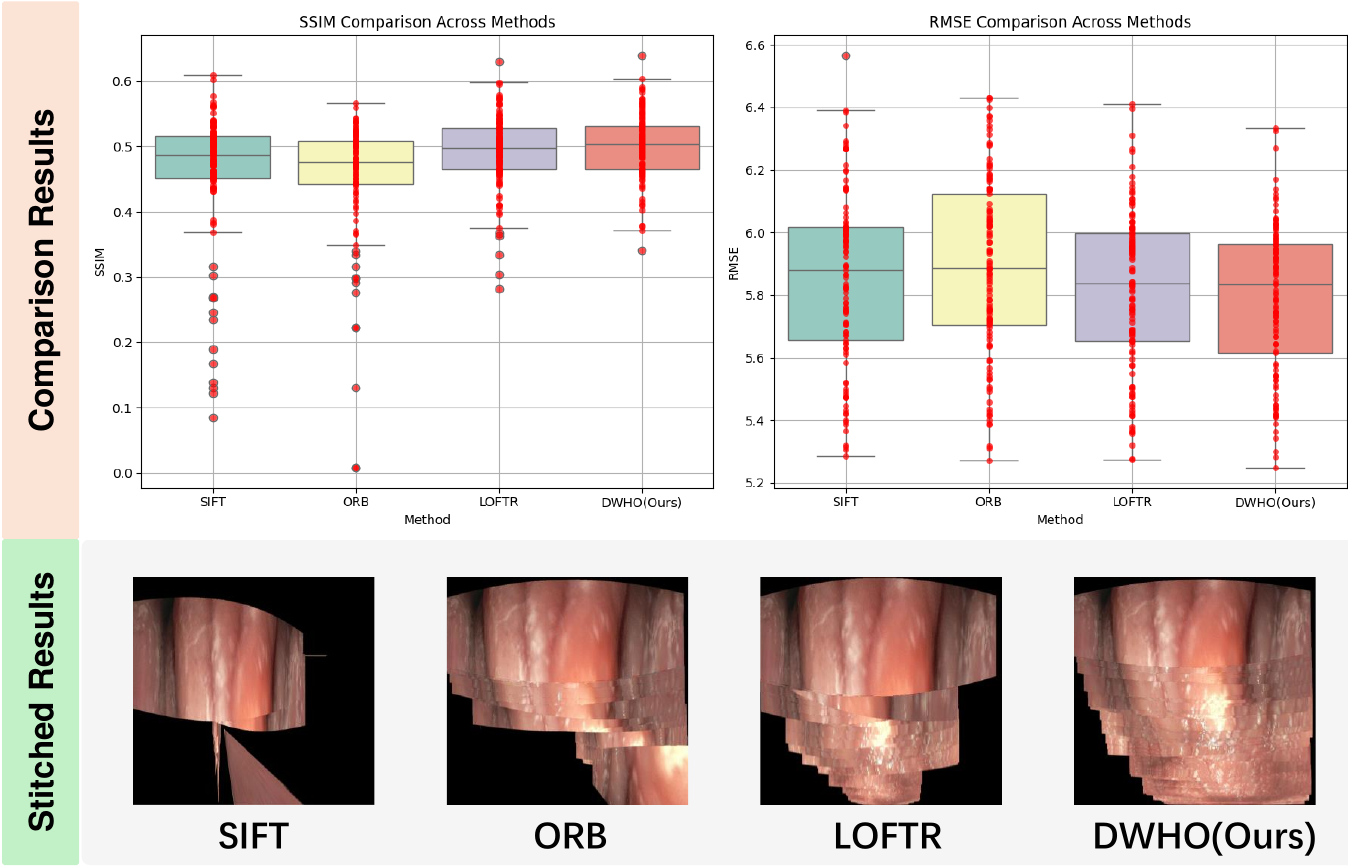}
\end{center}
\caption{Quantitative and Qualitative Comparison of Image Matching Methods. Top: Box plots comparing Structural Similarity Index (SSIM) and Root Mean Square Error (RMSE) across different methods (SIFT, ORB, LOFTR, DWHO).
Bottom: Visualized stitched results for each method, highlighting the performance in generating coherent and accurate image reconstructions.}
\label{fig:result}
\end{figure*}
\begin{table}[h]
\centering
\begin{tabular}{|l|c|c|c|c|}
\hline
\textbf{Method} & \textbf{Average SSIM} & \textbf{Average RMSE} & \textbf{P-value} \\ \hline

SIFT~\cite{lowe2004distinctive}            & 0.461              & 5.852              & $p < 0.05$                            \\ \hline
ORB~\cite{rublee2011orb}             & 0.454            & 5.888              & $p < 0.05$                           \\ \hline
LOFTR~\cite{2021arXiv210400680S}           & 0.492              & 5.815              & $p < 0.05$                            \\ \hline
DWHO(Ours)      & \textbf{0.502 }             & \textbf{5.776}              & Ref                                    \\ \hline

\end{tabular}
\caption{Comparative analysis of stitch methods with different Image Matching Methods, experimented with 300 frames from 5 esophagus videos individually}
\label{tab:comparison}
\end{table}

Table.~\ref{tab:comparison} compares and our proposed DWHO method with other method based on normal image matching methods—SIFT, ORB and LOFTR—using metrics from five video sequences. The SSIM values reflect the structural similarity between pairs of images, which are two adjacent frames from the video, during the stitching process, with DWHO achieving the highest average SSIM (0.502), indicating superior image quality. RMSE, calculated by comparing the images stitched using our method with manually annotated and stitched images, shows that DWHO also has the lowest RMSE (5.776), demonstrating better alignment accuracy. The difference between the reference (Ref.) method and benchmarks is statistically evaluated by Wilcoxon signed-rank test, resulting in a significant difference with $p < 0.05$.



Based on the quantitative analysis shown in Fig.\ref{fig:result}, our stitching process, which incorporates feature filtering and the DWHO algorithm, shows better performance in terms of stability and accuracy. The results indicate that our method achieves higher SSIM values, suggesting better structural integrity in the stitched images, and lower RMSE values, indicating fewer alignment errors. 

Moreover, the actual visualized stitching results are consistent with our quantitative comparison findings. The visual outcomes demonstrate that our method consistently produces stitched images with continuity and stability across multiple frames. These visual outcomes are supported by the quantitative analysis, which confirms that our DWHO algorithm significantly enhances structural similarity while minimizing errors.

\section{CONCLUSION}

\label{sec:conclusion}  
In this paper, we presented an advanced automatic image unfolding and stitching framework tailored for esophageal lining videos captured during endoscopy. By integrating feature matching techniques, including LoFTR, SIFT, and ORB, with our novel Density-Weighted Homography Optimization (DWHO) algorithm, we achieved to creating panoramic views of the esophagus. The experimental results, validated through RMSE analysis against manually refined images, demonstrated the method's consistency. This approach enhances the quality of endoscopic image analysis, providing clinicians with a comprehensive view.

\acknowledgments 
This research was supported by NIH R01DK135597(Huo), DoD HT9425-23-1-0003(HCY), NIH NIDDK DK56942(ABF).
This work was also supported by Vanderbilt Seed Success Grant, Vanderbilt Discovery Grant, and VISE Seed Grant. This
project was supported by The Leona M. and Harry B. Helmsley Charitable Trust grant G-1903-03793 and G-2103-05128.
This research was also supported by NIH grants R01EB033385, R01DK132338, REB017230, R01MH125931, and NSF
2040462. We extend gratitude to NVIDIA for their support by means of the NVIDIA hardware grant.
\bibliography{report} 
\bibliographystyle{spiebib} 

\end{document}